\documentclass{article}
\usepackage{ijcai21}

\usepackage{times}

\usepackage{soul}
\usepackage{url}
\usepackage[hidelinks]{hyperref}
\usepackage[utf8]{inputenc}
\usepackage[small]{caption}
\usepackage{graphicx}
\usepackage{amsmath,amssymb,amsfonts}
\usepackage{booktabs}
\usepackage[ruled]{algorithm2e}
\usepackage{algorithmic}
\usepackage{subfigure}
\usepackage{array}
\usepackage{makecell}
\usepackage{multirow}
\title{RH-Net: Improving Neural Relation Extraction via Reinforcement Learning and Hierarchical Relational Searching}

\author{
Jia-ning Wang
\affiliations
East China Normal University, Shanghai, China\\
\emails
lygwjn@gmail.com
}

\begin{document}

\maketitle

\begin{abstract}
Distant supervision (DS) aims to generate large-scale heuristic labeling corpus, which is widely used for neural relation extraction currently. However, it heavily suffers from noisy labeling and long-tail distributions problem. Many advanced approaches usually separately address two problems, which ignore their mutual interactions. In this paper, we propose a novel framework named RH-Net, which utilizes \textbf{R}einforcement learning and \textbf{H}ierarchical relational searching module to improve relation extraction. We leverage reinforcement learning to instruct the model to select high-quality instances. We then propose the hierarchical relational searching module to share the semantics from correlative instances between data-rich and data-poor classes. During the iterative process, the two modules keep interacting to alleviate the noisy and long-tail problem simultaneously. Extensive experiments on widely used NYT data set clearly show that our method significant improvements over state-of-the-art baselines.
\end{abstract}

\section{Introduction}
Relation extraction (RE) is a preliminary task in natural language processing (NLP), which aims to capture the relation between two target entities. Recently, RE based on conventional supervised learning has made a great success. However, it heavily relies on human annotations. 

\begin{figure}
\centerline{\includegraphics[height=42mm]{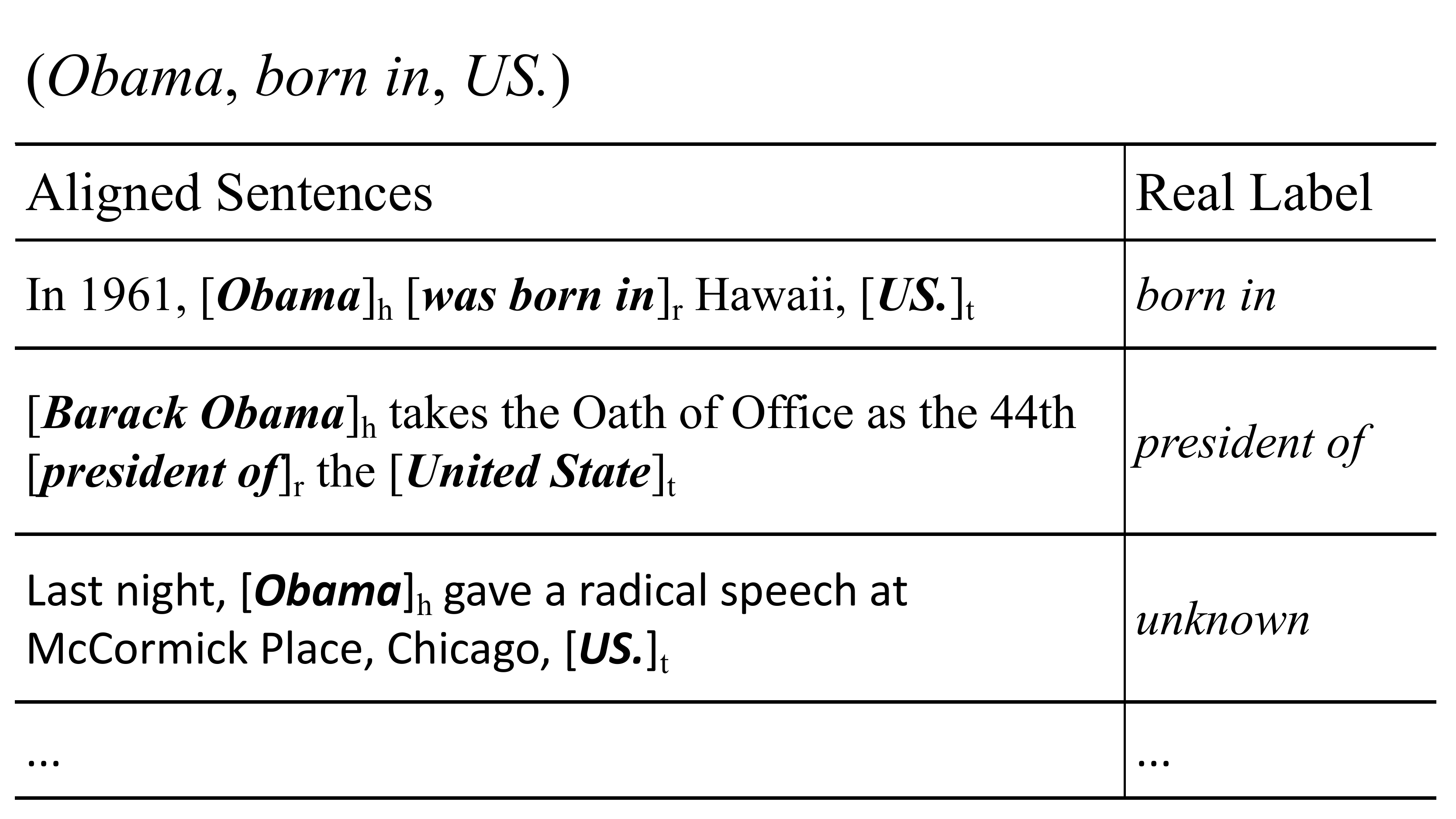}}
\caption{The example of sentence alignment from fact (\textit{Obama}, \textit{born in}, \textit{US.}) by distance supervision. It shows that only the first is correct labeling data and others are noise.}
\label{example}
\end{figure}

In order to obtain large-scale training corpus, distant supervision relation extraction (DSRE) \cite{mintz2009distant} was proposed to generate heuristic labeling data by aligning entity pairs in raw text. It assumes that if two target entities have a semantic relation in KG, all the raw text containing the two entities can be labeled as this relation class. However, this solution makes an over-strong assumption and inevitably brings in massive wrong labeling data. For example, as shown in Figure \ref{example}, given a fact (\textit{Obama}, \textit{born in}, \textit{US.}) from existing KG, DS will regard all sentences with two linked entities \textit{Obama} and \textit{US.} express the relation \textit{born in}. In consequence, only the first sentence is correct, but actually the second expresses the relation \textit{president of} while the third cannot find the pre-defined relation. Additionally, according to a series works \cite{li2019self-attention,xu2019connecting,han2018hierarchical,zhang2019long-tail}, DS always suffers from long-tail distribution problem. We analyze that there are two main reasons: 1) existing knowledge bases are far from completion and they contain the overlapping problem, 2) the number of noisy labeling sentences in some of the relation labels is larger than correct data, which causes the semantics or data insufficient. Inevitably, the first factor relies on the quality of KG, which is fixed before alignment with plain text. Therefore, we only devote to find the target solution corresponding to the second factor. By intuition, if there're a lot of noisy sentences under a relation class, fewer high-quality sentences can be sufficiently used to train the model, which results in the long-tail. In other words, To improve the relation extraction, the noisy labeling and long-tail problem should be considered simultaneously. Recently, most approaches have been presented to solve the noisy labeling problem \cite{hoffmann2011knowledge,Zeng2015Distant,Jat2018Improving,ji2017distant,Feng2018Reinforcement,Qin2018Robust,Qin2018DSGAN,zeng2018large} and long-tail problems \cite{Vashishth2018RESIDE,li2019self-attention,xu2019connecting,han2018hierarchical,zhang2019long-tail}. Despite the success and popularity of these methods, little works handle both two problems simultaneously, which ignore the mutual interactions.

In this paper, in order to jointly solve two problems, we propose a novel framework named RH-Net, which incorporates \textbf{R}einforcement learning and \textbf{H}ierarchical relational searching module. At first, we leverage reinforcement learning to select high-quality data. Concretely, given an original bag \footnote{Bag is the set of sentences which aligned with the same entity pair and labeled as the same relation class.}, the agent splits its into the correct set and noisy set, and we train the downstream module only on the correct set. This idea is motivated by the previous work \cite{Feng2018Reinforcement}, but the difference is that we enhance the agent by integrating pre-trained implicit relation information. For the second problem, we regard that the semantics of data-rich can be shared with similar data-poor relations. For example, the data-rich relation  \emph{/people/person/place\_of\_birth} in NYT corpus can represent a four-layers tree, from top to down are \emph{root}, \emph{/people}, \emph{/people/person} and \emph{/people/person/place\_of\_birth}, respectively, where \emph{root} is virtual node, \emph{/people} and \emph{/people/person} are sub-relations. When given a data-poor relation \emph{people/person/religion}, it can be integrated with related instances at the layer of \emph{root}, \emph{/people}, and \emph{/people/person}. In contrast to \cite{han2018hierarchical} and \cite{zhang2019long-tail}, we view RE as a tree search task from the root to the leaf node. During the search processing, we leverage the gating mechanism to save and combine the semantics of related instances at the current node, and calculate the score of each candidate child nodes and choose the maximum one. The two main components joint training at each iterative stage to capture the interactions. The contributions of this paper are as follows:
\begin{itemize}
    \item We are the first to transform the relation extraction into a tree search task. We propose the hierarchical relational searching strategy to share the correlated instance semantics at each node.
    
    \item We propose a novel framework RH-Net, which is capable of simultaneously solving the noisy labeling and long-tail problem. At the iterative training stage, our method takes advantage of the mutual interactions between them.
    
    \item Extensive experiments on the NYT data set demonstrate that if we consider both two problems, the proposed method outperforms state-of-the-art baselines.
\end{itemize}

\section{Related Work}

Distant supervision (DS) \cite{mintz2009distant} was proposed to automatically label large-scale corpus to overcome the time-consuming and human-intensive problem, which is one of the popular methods for semi-supervised relation extraction. However, it suffers from noisy labeling and long-tail distribution problems, which both cause by the over-strong heuristic assumption. 

Some recent researches solve these problems by employing multi-instance learning (MIL) for bag-level classification \cite{riedel2010modeling,hoffmann2011knowledge}. Inspired by MIL, \cite{Jat2018Improving,Lin2016Neural,Zeng2015Distant,ji2017distant} propose sentence-level attention, which can make the model focus on the high-quality sentence and reduce the influence of noise. Other works \cite{yuan2019cross-relation} denoise by extra bag-level attention to capture the correlation semantics between sentence and bag. Apart from MIL, recent studies have found a novel way to directly select correct data. \cite{Feng2018Reinforcement} is the first to utilize RL for RE. The instance selector (agent) is modeled as a binary-classifier, where 1 represents select action and 0 denotes remove action. The relation classifier is trained on the selected set and returns a reward through validation loss to the instance selector.  \cite{he2019improving} and \cite{Qin2018Robust} improved RL by using Q-network. In addition, \cite{Qin2018DSGAN,Han2018Denoising} leverage generative adversarial network (GAN) to filter noisy data by iterative training generator and discriminator. 

Additionally, some recent researches start to focus on the long-tail problem. For example, \cite{Vashishth2018RESIDE,li2019self-attention,xu2019connecting} utilize side information to realize semantics enhancement. \cite{beltagy2019combining} make data argumentation, such as entity type information, implicit or explicit relation-aware knowledge, etc. \cite{zhang2019long-tail,han2018hierarchical} leverage hierarchical attention to transfer data-rich information to data-poor class at the tail of the distribution, which succeeds in overcoming the long-tail problem without explicit external data. 

However, they only focus on one problem and ignore the communication between noisy labeling and long-tail distribution problem. To fill the gap in this part, we incorporate the reinforcement learning and hierarchical relational searching strategies to alleviate both noisy labeling and long-tail problem, and train them iteratively to make full use of the mutual interactions.

\begin{figure*}[htbp]
\centerline{\includegraphics[width=160mm]{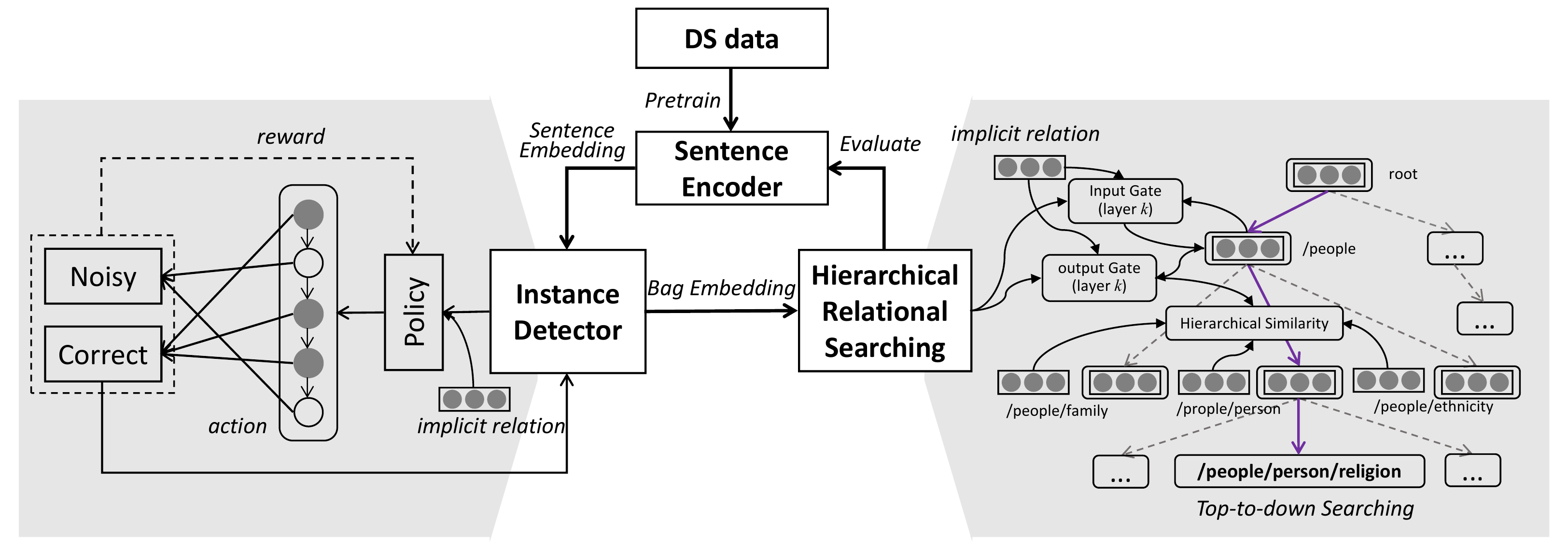}}
\caption{The architecture of our proposed RH-Net. The left is the instance detector, which aims to select high-quality data-driven by RL. The right is the hierarchical relational searching module, which views relation extraction as a top-to-down search processing.}
\label{pcnn_rl_HRS}
\end{figure*}%

\section{Methodology}

Figure \ref{pcnn_rl_HRS} illustrates our RH-Net, which consists of three main modules, including sentence encoder, instance detector, and hierarchical relational searching module.

% \subsection{Task and Notations}

% Given a KG $\mathcal{G=(E,R,F)}$ , where $\mathcal{E}$ represents the set of entities, $\mathcal{R}$ is the set of relations, $\mathcal{F\subseteq E\times R\times E}$ denotes the facts, where $(h, r, t)\in \mathcal{F}$, $r\in \mathcal{E}$ is the semantic relation between head entity $h\in\mathcal{E}$ and tail entity $t\in\mathcal{E}$. Given a DS dataset $T$, where $B\in T$ is the bag of sentences $S$ with corresponding aligned triple $(h_{B}, r_{B}, t_{B})$. The task aims to select the correct sentences from bag, and then to predict the semantic relation.

\subsection{Sentence Encoder}

We use PCNN \cite{Zeng2015Distant} to represent the sentence into a low-dimension vector. Given an input of sentence: $\mathbf{X} = [\mathbf{w}_1, \mathbf{w}_2, ..., \mathbf{w}_n]$, where $\mathbf{w}_i\in\mathbb{R}^{d_w+2 \times d_p}$ is the $i$th word vector consists of $d_w$-dimension pre-trained word embedding by GloVe \cite{pennington2014glove} and $d_p$dimension position embedding. We then use CNN with $K$ different $l$dimension filters to encode sentence by:
\begin{equation}
\mathbf{L} = CNN(\mathbf{X})
\label{eq1}
\end{equation}
where $\mathbf{L}=[\mathbf{L}^{(1)}, \mathbf{L}^{(2)}, ..., \mathbf{L}^{(K)}]\in\mathbb{R}^{K\times(n-l+1)}$. The piece-wise max pooling vector of the $j$-filter can be calculated by :
\begin{equation}
\mathbf{c}_j = [max(\mathbf{L}_{0:p_1}^{(j)}); max(\mathbf{L}_{p_1:p_2}^{(j)}); max(\mathbf{L}_{p_2:n}^{(j)})]
\label{eq2}
\end{equation}
where $p_1, p_2$ is the position of two entities $h_B,t_B$. $[\cdot;\cdot]$ is the concatenate operation.

% At last, we can output the sentence-level embedding represents $\mathbf{c}=[\mathbf{c}_1;\mathbf{c}_2;...;\mathbf{c}_K]\in\mathbf{R}^{d_c}$, where $d_c=3K$. We use cross entropy \cite{Zeng2015Distant} to train this module:
% \begin{equation}
% \mathcal{L}(\Pi) = -\frac{1}{N'}\sum_{i=1}^{N'}log p(r_i|S_i;\Pi) + \frac{\lambda_1}{2} ||\Pi||_2^2
% \label{eq3}
% \end{equation}
% where $\Pi$ denotes the parameters of sentence encoder, $\lambda_1$ is the L2 regularization parameters.

% defined as follow:
% \begin{equation}
% \mathcal{L}(\Pi) = -\frac{1}{N}\sum_{i=1}^{N}log p(r_i|S_i;\Pi)
% \label{eq2}
% \end{equation}
% where $\Pi$ denotes the parameters of sentence encoder, $N$ is the size of dataset.

\subsection{Instance Detector}

We propose the instance detector module based on RL to automatically split the original bag into the correct and noisy set. We follow \cite{Feng2018Reinforcement} to define the state, action, and reward function.

\textbf{State}. RL can be abstracted as the Markov decision process (MDP) of iterative interaction between the agent and the environment. In this paper, we regard the selection of a bag as an episode, and define the state embedding $\mathbf{s}_t$ consists of: 1) the average vector of selected sentences from correct set $\mathbf{\hat{x}}\in\mathbb{R}^{d_c}$, 2) the last state embedding $\mathbf{s}_{t-1}\in\mathbb{R}^{d_s}$, 3) the current sentence embedding $\mathbf{c}_t\in\mathbb{R}^{d_c}$ and 4) the implicit relation information $\mathbf{r}^*\in\mathbb{R}^{d_r}$. Obviously, different from \cite{Feng2018Reinforcement,he2019improving,Qin2018Robust}, we integrate the pre-trained implicit relation information to make enhancement. Formally:
\begin{equation}
\mathbf{s}_t = [tanh(\mathbf{W}_q[\mathbf{s}_{t-1}; \mathbf{c}_{t}; \mathbf{r}^*]); \mathbf{\hat{x}}_t]
\label{eq3}
\end{equation}
where $\mathbf{W}_q\in\mathbb{R}^{(d_s-d_c)\times(d_s+d_c+d_r)}$ is the trainable matrix. The implicit relation $\mathbf{r}^*=\mathbf{t-h}$, where $\mathbf{t,h}$ denotes the knowledge base embedding pre-trained by TransE \cite{Fan2014Transition}.

\textbf{Action}. At each time $t$, the instance detector takes an action to decide whether to select for correct set or remove for noisy set. It can be viewed as a binary-classifier refers to the policy $\pi_{\Theta} (a_i|\mathbf{s}_t)$:
\begin{equation}
\pi_{\Theta} (a_i|\mathbf{s}_t) = a_i\sigma (\mathbf{W}_p\mathbf{s}_t) + (1-a_i)(1-\sigma (\mathbf{W}_p\mathbf{s}_t))
\label{eq4}
\end{equation}
where $\sigma(\cdot)$ is the sigmoid function, $\mathbf{W}_p$ is the training matrix. $a_i\in\{0,1\}$ is the action space, where 1 denotes select action and 0 denotes remove action.

\textbf{Reward}. Generally, the reward function is used to evaluate the utility of agent. We follow \cite{Feng2018Reinforcement} to design a novel reward function. We assume that the model has a terminal reward when it finishes all the selection:
% \begin{equation}
% \begin{aligned}
% & R(B) = \frac{M_{cre}}{M(M_{cre} + \gamma)}[\sum_{S_i\in B_{col}}p(r_B|S_i) + \gamma]\\
% &+ \frac{M_{noi}}{M} \{1-\frac{1}{M_{noi} + \gamma}[\sum_{S_j\in B_{noi}}p(r_B|S_j) + \gamma]\}
% \label{eq6}
% \end{aligned}
% \end{equation}

\begin{equation}
\begin{aligned}
R(B) = \frac{1}{M_{cre}}\sum_{S_j\in B_{cre}}\text{log}p(r_B|S_j)
\label{eq5}
\end{aligned}
\end{equation}
where $M_{cre}$ is the number of sentences in correct set $B_{cre}$. Note that, we accumulate the log-aware probability of each sentence $p(r_B|S_i)$ in each subset to represent the occurrence probability of ground truth. We train this module by REINFORCE algorithm \cite{williams1992simple} and following the same settings by \cite{Feng2018Reinforcement}. At last, we obtain the correct set $B_{cre}$ and corresponding bag-level embedding $\mathbf{\hat{x}}$.

% \begin{figure}
% \centerline{\includegraphics[width=80mm]{HRS.pdf}}
% \caption{The hierarchical tree structure}
% \label{HRS}
% \end{figure}

\subsection{Hierarchical Relational Searching (HRS)}

After denoising, we propose this module to extract the semantics relation. We introduce this module from three aspects:

\textbf{The construction of hierarchical relation tree}. Given an original relation label $r\in\mathcal{R}$, it can be represented as a path from root (layer4) to the leaf (layer1), and the node at layer $k$ denotes $r_k$ ($k\in\{1,2,3,4\}$). In addition, we suppose that $r'_k$ is the sibling node of $r_k$. The child nodes set of $r_k, r'_k$ represents $\mathcal{N}_k(r)$ and $\overline{\mathcal{N}_k(r)}$, respectively, where $\overline{\mathcal{N}_k(r)}$ is the complement set of $\mathcal{N}_k(r)$. Therefore, we have $r_{k-1}, r'_{k-1}\in\mathcal{N}_k(r)$, $r''_{k-1}\notin\mathcal{N}_k(r)$, where $r_{k-1}$ is the true path at layer $k-1$, $r'_{k-1}$ is the negative but share the same parent node, $r''_{k-1}$ is the negative but not share the same parent node.

As shown in Figure \ref{pcnn_rl_HRS}, each node consists of sub-relation embedding and memory cell embedding. The original relation embedding at layer 1 is pre-trained by TransE \cite{Fan2014Transition}, and then we recursively calculate the embedding of each sub-relation from layer 2 to 4. Formally:
\begin{equation}
\mathbf{r}_k = \frac{1}{|\mathcal{N}_k(r)|}\sum_{r_{k-1}\in\mathcal{N}_k(r)}\mathbf{r}_{k-1}
\label{eq6}
\end{equation}
where $\mathbf{r}_k\in\mathbb{R}^{d_r}$ is the sub-relation embedding of node $r_k$, $|\mathcal{N}_k(r)|$ denotes the number of child nodes. The memory cell embedding aims to preserve the semantics of instances\footnote{Instance is referred to a bag in this module}, which initialed as zeros vector $\mathbf{C}_k{(r)}=\mathbf{0}\in\mathbb{R}^{d_{cell}}$.

\textbf{The Top-to-down search processing with gating mechanism}.
Given a bag $B$ and the bag-level embedding $\mathbf{\hat{x}}$ which outputs from instance detector. HRS aims to search a path from the root to the leaf node, which can be also viewed as a multi-branch classification between two adjacent layers. Specifically, we first obtain the fusion of implicit relation information and bag-level embedding $\mathbf{G} = tanh(\mathbf{W}_G[\mathbf{\hat{x}};\mathbf{r}^{*}] + \mathbf{b}_G)$, where $\mathbf{r}^{*}=\mathbf{t-h}$ denotes implicit relation, $\mathbf{W}_G\in\mathbb{R}^{d_{cell}\times d_c}$ and $\mathbf{b}_G\in\mathbb{R}^{d_{cell}}$ is the trainable parameters. $\mathbf{G}\in\mathbb{R}^{d_{cell}}$ is the fusion information of one bag. Suppose that the bag at node $r_k$, inspired by GRUs and LSTMs, we use an input gate $i_k(r)$ to selective save this fusion information to update the memory cell $\mathbf{C}_k^{old}(r)$ to $\mathbf{C}_k^{new}(r)$ at the node $r_k$:
\begin{equation}
i_k(r) = \sigma(\mathbf{W}_{i,k}[\mathbf{\hat{x}};\mathbf{C}_k(r)] + \mathbf{b}_{i,k})
\label{eq7}
\end{equation}
\begin{equation}
\mathbf{C}_k^{new}(r) = i_k(r)\cdot\mathbf{G} + (1 - i_k(r))\cdot\mathbf{C}_k^{old}(r)
\label{eq8}
\end{equation}
We then use an output gate $o_k(r)$ to extract the mixed semantics from memory cell at the node $r_k$:
\begin{equation}
o_k(r) = \sigma(\mathbf{W}_{o,k}[\mathbf{\hat{x}};\mathbf{C}_k^{new}(r)] + \mathbf{b}_{o,k})
\label{eq9}
\end{equation}
\begin{equation}
\mathbf{Z}_k(r) = o_k(r)\cdot\mathbf{C}_k^{new}(r) + (1 - o_k(r))\cdot\mathbf{G}
\label{eq10}
\end{equation}
where $\mathbf{W}_{i,k}, \mathbf{W}_{o,k}, \mathbf{b}_{i,k}, \mathbf{b}_{o,k}$ are the trainable matrices and bias at the layer $k$, $\sigma(\cdot)$ is the sigmoid function, $[\cdot ;\cdot]$ is the concatenate operation. $\mathbf{Z}_k(r)$ is the mixed semantics of bag $B$ at the node $r_k$, we can calculate the score of each next branch to child node $r_{k-1}\in\mathcal{N}_k(r)$, and choose the maximum one $r^{*}_{k-1}$ as the next node.
\begin{equation}
f(\mathbf{Z}_k(r), \mathbf{r}_{k-1}) = softmax(\mathbf{Z}_k(r)\mathbf{W}_{f,k}\mathbf{r}_{k-1}^{\mathbb{T}})
\label{eq11}
\end{equation}
\begin{equation}
r^{*}_{k-1} = \mathop{\arg\max}\limits_{r_{k-1}}f(\mathbf{Z}_k(r), \mathbf{r}_{k-1})
\label{eq12}
\end{equation}
where $\mathbf{W}_{f,k}$ is the matrix of score function $f(\cdot)$ at layer $k$.

\textbf{The hierarchical weighted ranking loss}. We find that cross-entropy and hierarchical metric learning \cite{verma2012learning} cannot be directly used, because 1) each node has a different number of branches, 2) when training one node, parameters of the other nodes will change which results in local optimum or divergence, 3) each layer or node has different influence degree on the loss.

Thus, we propose hierarchical weighted ranking loss. Firstly, we use the ranking loss to replace conventional cross-entropy loss. In other words, we use the opposite of the score as the loss function, and train to maximize the score of the correct path. Secondly, we additionally perform negative sampling from two aspects: 1) $r'_{k-1}\in\mathcal{N}_k(r)$ and 2) $r''_{k-1}\in\overline{\mathcal{N}_k(r)}$. Therefore, the loss function at the layer $k(k\in\{2,3,4\})$ can be defined as follows:
\begin{equation}
\begin{aligned}
& \mathcal{L}_k(B, r) = \\
& \sum_{r'_{k-1}\in\mathcal{N}_k(r)}||f(\mathbf{Z}_k(r), \mathbf{r'}_{k-1})
+ \mu - f(\mathbf{Z}_k(r), \mathbf{r}_{k-1})||_{+}\\
& + \sum_{r''_{k-1}\in\overline{\mathcal{N}_k(r)}}||f(\mathbf{Z}_k(r), \mathbf{r''}_{k-1})
+ \mu - f(\mathbf{Z}_k(r), \mathbf{r}_{k-1})||_{+}
\label{eq13}
\end{aligned}
\end{equation}
where $\mu\in[0,1]$ is the margin hyper-parameter, $||\cdot||_{+}$ is the hinge function. 

At last, We heuristically design weighted value by considering that the model should pay more attention to the node which is near to root or has too many child nodes:
\begin{equation}
\alpha_k(r) = \frac{|\mathcal{N}_k(r)|+k-1}{\sum_{j=2}^{4}(|\mathcal{N}_j(r)|+j-1)}
\label{eq14}
\end{equation}
The final loss of this module defined as:
\begin{equation}
\mathcal{L}(B, r) = \sum_{k=2}^{4}\alpha_k(r)\mathcal{L}_k(B,r)
\label{eq15}
\end{equation}

% \begin{algorithm}[t]
% \caption{RH-Net}
% \label{alg}
% \KwIn{DS training data $\mathbf{B}$, pre-trained sentence encoder $\Pi$ and instance detector $\Theta$, initialized HRS $\Phi$, iteration number $L$, a small $\tau$.}
% \KwOut{Parameter $\Pi'$, $\Theta'$ and $\Phi'$\;}
% Initialize sentence encoder $\Pi' = \Pi$, instance detector $\Theta' = \Theta$ and HRS $\Phi'=\Phi$ \;
% \For{iteration l=\text{1} to L}
% {
%   Shuffle training data $\mathbf{B}$, correct set $\mathbf{B}_{cre} = \emptyset$\;
%   \ForEach{$B_i \in \mathbf{B}$}
%   {
%     For each sentence, obtain the state and sample action by Equal \ref{eq4} and \ref{eq5} with $\Theta'$\;
%     Comput delayed reward by Equal \ref{eq6}\;
%     Update the parameter $\Theta$\;
%   }
%   $\Theta'=\tau\Theta + (1-\tau)\Theta'$\;
%   \ForEach{$B_i \in \mathbf{B}$}
%   {
%     Obtain correct set $\hat{B}_{i}$ of $B_i$ with $\Theta'$\;
%     Add $\hat{B}_{i}$ into $\mathbf{B}_{cre}$\;
%   }
  
%   \ForEach{$B_i\in \mathbf{B}_{cre}$}
%   {
%     Obtain the bag-level embedding of $B_i$\;
%     \While{not at leaf (layer $k > 1$)}
%     {
%         Save semantics by Equal \ref{eq9} and \ref{eq10}\;
%         Obtain mixed semantics $\mathbf{Z}_k$ by Equal \ref{eq11} and \ref{eq12}\;
%         Search for next node by Equal \ref{eq14}\;
%     }
%     Calculate the loss of the HRS by Equal \ref{eq17} and update parameter $\Phi$\;
%   }
%   Update the sentence encoder $\Pi$\;
%   $\Phi'=\tau\Phi + (1-\tau)\Phi'$\;
%   $\Pi'=\tau\Pi + (1-\tau)\Pi'$\;
% }
% \end{algorithm}

\subsection{The training strategy}

In order to accelerate the training and make the model stable, we follow previous work \cite{Feng2018Reinforcement,Qin2018Robust} to pre-train at first. Concretely, we pre-train sentence encoder to obtain sentence-level embedding, and then pre-train the instance detector by computing the reward function. The pre-train stage of RL is crucial for our experiment. 

At the joint training stage, the parameters of the sentence encoder are fixed. We first train the instance detector and obtain the correct set. Then, we train the hierarchical relational searching module on the correct set. Specifically, the instance detector selects a few high-quality sentences under the guidance of the reward function, and then feeds into the hierarchical relational searching module, their semantics can be associated with other samples via a gating mechanism. The top-to-down search processing aims to simulate the prediction of long-tail relations, and the result can then be used as a new reward function to update the first module. The whole process is executed iteratively. In this manner, our proposed RH-Net fully considers the interaction between the two problems, so as to further improve the performance of relation extraction.

\section{Experiment}
\subsection{Dataset and Evaluation Metrics}

%https://github.com/ZhixiuYe/Intra-Bag-and-Inter-Bag-Attentions%
We evaluate the proposed framework on widely used Distant Supervision (DS) dataset NYT \cite{riedel2010modeling}. The dataset \footnote{The NYT dataset can be download from http://iesl.cs.umass.edu/riedel/ecml/} has 52 semantic relations and a special NA label which means that no relation between entity pair. The training set contains 522611 sentences, 281270 entity pairs, and 18252 relational facts. The testing set contains 172448 sentences, 96678 entity pairs, and 1950 relational facts.

%https://github.com/thunlp/OpenNRE%
To fairly compare with some baselines\footnote{https://github.com/thunlp/OpenNRE}, we follow \cite{Lin2016Neural} to evaluate our method in the held-out evaluation and manual evaluation. The held-out evaluation aims to compare the predicted relational fact from the test data with the facts in Freebase, but it does not consider the efficiency of predicting NA class. The manual evaluation is performed to avoid the influence of the noisy testing data by manually checking the efficiency. We select precision-recall (P-R) curve, P@N, and Hits@K metrics to report the results of the experiment.

\subsection{Experiment Settings}

In sentence encoder, we use the same hyper-parameters as previous works \cite{Zeng2015Distant}. The word embedding size $d_w=50$, The position embedding size $d_p=5$. The filters $K=230$ and the window size $l$ is set to 3. The implicit relation and memory cell embedding dimension $d_r=d_{cell}=50$. The batch size is 64. The learning rate was 0.02, 0.01 at the pre-training and joint training stage, respectively. We employ a dropout strategy with a probability of 0.5. The small constant $\mu=0.5$. We pre-train sentence encoder and instance detector for 5 epochs. The joint training iteration number $L$ is 30. We apply Adam \cite{kingma2014adam} method to optimize parameters both at the pre-training and iterative training stage.

\begin{figure}[t]
\centerline{\includegraphics[width=76mm]{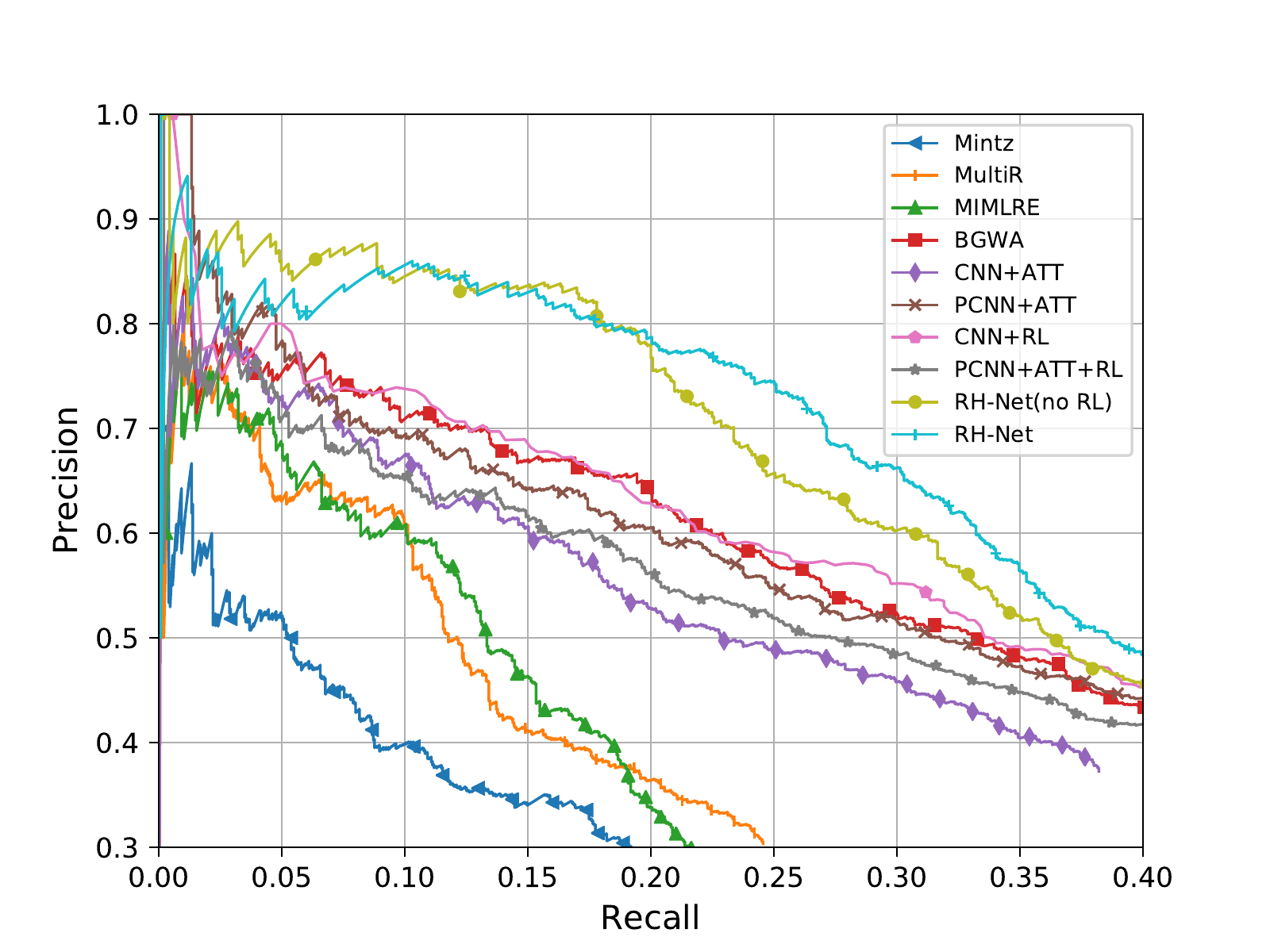}}
\caption{Comparison with previous baselines.}
\label{main result}
\end{figure}%

\subsection{Comparison Methods}

We use held-out evaluation to compare our model \textbf{RH-Net} with several baselines. 

\begin{itemize}
    \item \textbf{Feature-based methods} utilize traditional feature engineering and machine learning, such as \textbf{Mintz} \cite{mintz2009distant}, \textbf{MultiR} \cite{hoffmann2011knowledge} and \textbf{MIML} \cite{Surdeanu2012Multi}. \textbf{Mintz} is a traditional method for DSRE via human designed features and multi-class logistic regression. \textbf{MultiR} leverages MIL to reduce the noise and handle the overlapping problem by proposed probabilistic graphical module. \textbf{MIML} utilizes multi-instance multi-label method for extracting semantics relation between two entities.
    
    \item \textbf{Soft-strategy methods} leverage sentence-level attention mechanism to reduce the influence of noisy, including \textbf{BGW-A} \cite{Jat2018Improving}, \textbf{CNN +ATT} and \textbf{PCNN +ATT} \cite{Lin2016Neural}. \textbf{BGWA} is a bidirectional GRU based RE model with piecewise max pooling. \textbf{CNN +ATT} is a basic CNN module with sentence-level attention mechanism. \textbf{PCNN +ATT} combines the sentence-level attention with PCNN to capture structure information between two entities.
    
    \item  \textbf{Hard-strategy methods} aim to filter noisy before RE, consisting of \textbf{CNN +RL}  \cite{Feng2018Reinforcement} and \textbf{PCNN +ATT +RL} \cite{Qin2018Robust}. \textbf{CNN +RL} is a novel method to reduce noisy labeling data by RL, and achieves rewards from CNN to evaluate the RL. \textbf{PCNN +ATT +RL} also introduce the RL, but it redistributes noisy sentences into negative examples.
\end{itemize}
Note that, we find a new work \cite{he2019improving}, which is also improved by reinforcement learning belongs to hard-strategy methods, but they do not provide the code or corresponding experimental P-R curve raw result data. Therefore, we ignore this comparison.

\subsection{Main Results}

As shown in Figure \ref{main result}, we use the P-R curve to make a comparison without NA label \footnote{We follow previous works to only preserve the precision and recall value of no-NA labels, because that the prediction of NA is useless.}, where the x-axis denotes the recall and y-axis denotes the precision. The main results indicate that 1) both soft-strategy and hard-strategy based on deep learning methods outperform the feature-based methods, it means that the representation and generalization of traditional feature engineering unable to improve the performance. 2) The performance of the CNN-based method is worse than the PCNN-based, this is due to the factor that CNN ignores the entity structure information, while other methods consist of piecewise max-pooling can make reliable promotion. 3) We also find our proposed RH-Net outperforms all other baselines by a large margin, which demonstrates that the successful improvement of considering two problems and integrating two main solution modules to joint train. 4) We also train our method without reinforcement learning (denotes RH-Net(no RL)), we find that the result will be slightly worse than RH-Net with RL. It indicates that the noisy labeling data can disturb the model to learn semantics relations.

\begin{table}
\centering
\renewcommand\tabcolsep{3.0pt}
\begin{tabular}[85mm]{cccccccc}
\hline
\multicolumn{2}{c}{\multirow{2}{*}{\begin{tabular}[c]{@{}c@{}}Training Instances\\ Hits@K(Marco)\end{tabular}}} & \multicolumn{3}{c}{\textless{}100} & \multicolumn{3}{c}{\textless{}200} \\
\multicolumn{2}{c}{} & 10 & 15 & 20 & 10 & 15 & 20 \\ \hline
\multirow{5}{*}{CNN} & +ATT & \textless{}5.0 & \textless{}5.0 & 18.5 & \textless{}5.0 & 16.2 & 33.3 \\
 & +HATT & 5.6 & 31.5 & 57.4 & 22.7 & 43.9 & 65.1 \\
 & +KATT & 9.1 & 41.3 & 58.5 & 23.3 & 44.1 & 65.4 \\
 & \textbf{+HRS} & 9.5 & 40.2 & 59.6 & 23.9 & 47.0 & 66.6 \\
 & \textbf{+RL+HRS} & \textbf{11.3} & \textbf{41.5} & \textbf{60.1} & \textbf{25.0} & \textbf{47.1} & \textbf{66.9} \\ \hline
\multirow{5}{*}{PCNN} & +ATT & \textless{}5.0 & 7.4 & 40.7 & 17.2 & 24.2 & 51.5 \\
 & +HATT & 29.6 & 51.9 & 61.1 & 41.4 & 60.6 & 68.2 \\
 & +KATT & 35.3 & 62.4 & 65.1 & 43.2 & 61.3 & 69.2 \\
 & \textbf{HRS} & \textbf{36.8} & 64.0 & 68.8 & \textbf{44.8} & 62.0 & 71.5 \\
 & \textbf{+RL+HRS} & 36.6 & \textbf{64.1} & \textbf{68.9} & 44.5 & \textbf{62.3} & \textbf{71.7} \\ \hline
\end{tabular}
\caption{The marco accuracy of Hits@K on long-tail relations. \textbf{+RL} means using reinforcement learning for denoising, \textbf{+HRS} means using hierarchical relational searching module.}
\label{longtail_result}
\end{table}

\begin{table}
\centering
\setlength\tabcolsep{7pt}
\begin{tabular}{c|cccc}
\hline
\multirow{2}{*}{Methods} & \multicolumn{4}{c}{Precision} \\
 & P@100 & P@200 & P@300 & Avg. \\
\hline
\textbf{RH-Net} & 90.00 & 84.50 & 79.33 & 84.61 \\
\textbf{w/o IR} & 88.11 & 81.93 & 77.50 & 82.51 \\
\textbf{w/o GM} & 82.19 & 74.00 & 63.33 & 73.17 \\
\textbf{w/o WL} & 83.26 & 77.33 & 71.35 & 77.31 \\
\hline
\end{tabular}
\caption{The ablation results of RH-Net on NYT data set.}
\label{ablation_result}
\end{table}

\begin{table*}
\centering
\small
\setlength\tabcolsep{3pt}
\begin{tabular}{p{8.1cm}|m{3.7cm}|m{3.7cm}|m{1.2cm}}
\hline
\makecell[c]{Sentences} & \makecell[c]{Original label} & \makecell[c]{Predicted label}  & \makecell[c]{Is noise?}  \\
\hline
Kuhn and his wife luisa relocated to \textbf{ponte vedra} beach , \textbf{florida} in 1990 , ... & \makecell[c]{/location/location/contains} & \makecell[c]{/location/location/contains} & \makecell[c]{No} \\
\hline
a former military ruler , \textbf{muhammadu buhari} , also a northern \textbf{muslim} , is a leading candidate , ... & \makecell[c]{/people/person/religion} & \makecell[c]{/people/person/religion} & \makecell[c]{No} \\
\hline
... the american rights to \textbf{jonathan littell}’s novel les bienveillantes, which became a publishing sensation in \textbf{france}, have been sold to harpercollins, ... & \makecell[c]{/people/person/nationality} & \makecell[c]{NA} & \makecell[c]{Yes} \\
\hline
... the annual meeting morphed into a three and a half hour celebration of \textbf{sanford i. weill} , \textbf{citigroup} 's departing chairman & \makecell[c]{/business/company/advisors} & \makecell[c]{/business/company/founders} & \makecell[c]{No} \\
\hline
\end{tabular}
\caption{Some sentences for case study.}
\label{case_result}
\end{table*}

\section{Analyses}

\subsection{The results for long-tail relations}

We also demonstrate the improvements for long-tail relations. We choose three attention-based models \textbf{+ATT} \cite{Lin2016Neural}, \textbf{+HATT} \cite{han2018hierarchical} and \textbf{+KATT} \cite{zhang2019long-tail}:
\begin{itemize}
    \item \textbf{+ATT} is the traditional sentence-level attention mechanism over instances, such as CNN+ATT and PCNN+ATT\cite{Lin2016Neural}.
    
    \item \textbf{+HATT}  is the hierarchical attention method over the instances, the difference is that it considers the hierarchical structure of semantic relation.
    
    \item \textbf{+KATT} is also an attention-based method, which utilizes knowledge base embedding (KBE) and graph neural network (GNN) to represent the hierarchical relational label.
    
\end{itemize}
To make a fair comparison, we follow the same evaluation strategy by them. Specifically, we obtain a subset from testing data in which all the relations have fewer than 100 or 200 instances, we leverage the macro Hits@K metric, which means that the accuracy of the golden relation in the top K candidate relations recommended by our model. In this experiment, we select K from $\{10,15,20\}$. 

Note that, the \textbf{PCNN+RL+HRS} denotes our proposed RH-Net, where \textbf{+RL} means using reinforcement learning and \textbf{+HRS} means using hierarchical relational searching. \textbf{CNN+RL+HRS} is the same as RH-Net, except for the encoder is CNN. As shown in Table \ref{longtail_result}, it illustrates that: 1) The PCNN-based encoder is better than CNN, which indicates that the piecewise information is also useful for long-tail prediction. 2) HRS module with both CNN and PCNN outperforms previous works, it verifies that the hierarchical tree processing is really better than simple attention. 3) If we use RL to filter the noisy data before relation extraction, despite obtaining a bit of improvement, it is still hard to extract the long-tail relations because of the reduction of data.

\subsection{Ablation Study}

We perform ablation experiments to validate the contributions of different components of our models. We report the P@N metric, which denotes the top N of precision. Specifically, we evaluate all the testing instances and achieve the corresponding sorted precision value at layer 1, and then we choose the N-th value as P@N. We remove the following settings: 
\begin{itemize}
    
    \item \textbf{w/o IR} is the method without implicit relation in instance detector or HRS.
    
    \item \textbf{w/o GM} is the method without gating mechanism, which calculates the score function by only the semantics of instance itself.
    
    \item \textbf{w/o WL} is the method without the weighted influence of different layers or nodes, it means that we replace the Equal \ref{eq15} with a simple average operation.
\end{itemize}

As shown in Table \ref{ablation_result}, we find that if we remove one of these components, the performance of both RH-Net will be worse. Specifically, 1) if we ignore the implicit relation, the average of P@N will reduce by 2.10\%, owing to the agent missing some semantics of implicit relation information. 2) when we remove the gating mechanism, the average of P@N metric will reduce by more than 10\%, it illustrates that the gating mechanism is vital for sharing the knowledge between related instances, it is also an important part of our method to deal with the long-tail problem. 3) when we ignore the weighted sum loss, the average of P@N value will be reduced by 7.30\%, which means that the weighted sum loss of different layers or nodes makes positive contributions to stable training.

\subsection{Case Study}

We further present some sentences in Table \ref{case_result} for the case study. The text in bold represents the entity. The first two sentences which belong to a long-tail class, successfully selected by the instance detector and predicted by HRS. The third noisy sentence is removed for the noisy set and directly predicted as NA. Our method makes the wrong prediction on the last sentence, we analyze that the sample number of \textit{/business/company/advisors} is too small to predict the third layer, but our HRS still performs well in the first two layers.

\section{Conclusion}

In this paper, we propose a novel framework to alleviate both noisy labeling and long-tail problem. We apply RL to select the correct data and improve the RL by implicit relation information and a novel reward function that consider the contributions of both correct and noisy data. For the long-tail problem, we newly transform the relation extraction into a tree searching task and share the semantics of related instances between data-rich classes at the head of distribution and data-poor classes at the tail. We also provide a hierarchical weighted loss function to train this module. Extensive experimental results on the NYT dataset show that our method outperforms state-of-the-art baselines. In the future, we will pay attention to the overlapping problem. We also decide to apply this proposed framework to the few-shot RE task.

% \section*{Acknowledgments}
% We would like to thank the anonymous reviewers for their
% insightful comments and suggestions.

\bibliographystyle{named}
\bibliography{ijcai21}

\end{document}